\documentclass{article}
\usepackage{arxiv}
\usepackage{color,soul}

\usepackage[utf8]{inputenc} 
\usepackage[T1]{fontenc}    
\usepackage{hyperref}       
\usepackage{url}            
\usepackage{booktabs}       
\usepackage{amsfonts}       
\usepackage{nicefrac}       
\usepackage{microtype}      
\usepackage{float}
\usepackage{amsmath,amsfonts,amssymb,graphicx}
\usepackage{natbib}
\usepackage{comment}

\usepackage{caption}
\usepackage{subcaption}

\title{Improving the Accuracy and Robustness of CNNs Using a Deep CCA Neural Data Regularizer}

\author{%
Cassidy Pirlot \\
    Departments of Computing Science and Psychology\\
  University of Alberta, Edmonton, Canada \\
  \texttt{pirlot@ualberta.ca} \\
  \And
     Richard C. Gerum\\
  Department of Physics and Astronomy\\
  York University University\\
  Toronto, ON, Canada \\
  \texttt{richard.gerum@protonmail.com} \\
  \AND
   Cory Efird\\
  Departments of Computing Science and Psychology\\
  University of Alberta, Edmonton, Canada \\
  \texttt{efird@ualberta.ca} \\
  \AND
    Joel Zylberberg\\
  Department of Physics and Astronomy\\
  York University University\\
  Toronto, ON, Canada \\
  \texttt{joelzy@yorku.ca} \\
  \And
    Alona Fyshe \\
  Departments of Computing Science and Psychology\\
  University of Alberta, Edmonton, Canada \\
  \texttt{alona@ualberta.ca} \\
}

\begin{document}
\maketitle

\begin{abstract}
As convolutional neural networks (CNNs) become more accurate at object recognition, their representations become more similar to the primate visual system. This finding has inspired us and other researchers to ask if the implication also runs the other way: If CNN representations become more brain-like, does the network become more accurate? Previous attempts to address this question showed very modest gains in accuracy, owing in part to limitations of the regularization method. To overcome these limitations, we developed a new neural data regularizer for CNNs that uses Deep Canonical Correlation Analysis (DCCA) to optimize the resemblance of the CNN's image representations to that of the monkey visual cortex. Using this new neural data regularizer, we see much larger performance gains in both classification accuracy and within-super-class accuracy, as compared to the previous state-of-the-art neural data regularizers. These networks are also more robust to adversarial attacks than their unregularized counterparts. Together, these results  confirm  that neural data regularization can push CNN performance higher, and introduces a new method that obtains a larger performance boost.
\end{abstract}

\section{Introduction}
Convolutional neural networks (CNNs) have revolutionized computer vision, enabling object recognition performance comparable to that of humans~\citep{He2015}. While the architecture of CNNs was motivated by that of the primate \footnote{The taxonomic order that includes humans, monkeys, and apes.} visual system~\citep{lecun2015deep}, CNNs clearly do not exactly mimic the \emph{function} of the primate visual system. For example, CNNs are vulnerable to adversarial attacks~\citep{kurakin2016adversarial,goodfellow2014explaining}, in which small perturbations are added to an image's pixel values (often imperceptible to human observers) cause trained CNNs to change their object category predictions. Owing to this mismatch, there is strong motivation amongst computational neuroscientists and machine learning researchers, to generate CNNs that process images more similarly to the way our brains do~\citep{Federer2020,nassar20201,safarani2021towards}. In this paper, we introduce a new technique for achieving this goal, based on Deep Canonical Correlation Analysis (Deep CCA, DCCA~\citep{AndrewDCCA}). In this paper our contributions are as follows: Firstly, compared to unregularized networks and to other previously published neural data (ND) regularizers, our DCCA approach produces more accurate networks. Secondly, when the network misclassifies an image, it is more likely to be a reasonable mistake (within the same super-class). Thirdly, we show that our ND regularized networks are more robust to adversarial attacks of moderate strength. These results point to the promise of pushing CNN representations to be more brain-like.

\section{Background}
Previous work has shown that there is a correlation between the image representations in the primate visual system and the representations learned by CNNs, and that this correlation grows as the CNNs become more accurate at image categorization.  But does the implication run the other way?  Do CNNs that have more brain-like representations also become more accurate?
Previous research has used two main approaches to create CNNs with representations that are more similar to the brain's:

\begin{itemize}
    \item In the first approach, neural data in response to images is measured.  These neural response patterns are then used as a regularizer to constrain the image representations within the CNN trained for an object recognition task~\citep{Federer2020,safarani2021towards}. This procedure forces the CNN to learn the object recognition task while also forming more brain-like image representations. 

Prior work attempted to achieve this goal in two different ways: a) by training a read out of the intermediate layer activations in the CNN to predict neural firing rates when a monkey views the same images that were input to the network~\citep{safarani2021towards}; and b) by optimizing the match between the representation similarity matrix~\citep{kriegeskorte2008representational} of an intermediate CNN layer and that obtained from the neural recordings~\citep{Federer2020}. These two closely-related procedures were shown to improve robustness of the CNN to image distortions~\citep{safarani2021towards}, to slightly improve the CNN's accuracy~\citep{Federer2020}, and to improve robustness to label corruption during training~\citep{Federer2020}, when compared to CNNs that did not use neural data in their training procedure. 

\item The second approach used by previous research was to identify spectral properties of neural representations (namely, the $1/n$ eigenspectrum of the covariance matrix of neural firing rates in~\citet{Stringer2019}), and then to regularize CNNs to match those spectral properties while they were being trained for object recognition~\citep{nassar20201}. This approach led to CNNs that demonstrated improved robustness to adversarial attacks when compared to CNNs that were not regularized to match the spectral properties of neural representations.
\end{itemize}

Altogether, the previous work demonstrates tantalizing hints that regularizers that force CNNs to more closely mimic the brain can improve their object recognition performance and robustness. Despite these successes, there are several limitations to the previous works that we aim to overcome here. First, the improvements in robustness to adversarial attacks reported by~\citet{nassar20201} were limited to categorizing MNIST images of hand-written digits, leaving it unclear whether the same effects would be obtained on tasks involving more complex natural images. There were also no indications that the spectral regularizer of~\citet{nassar20201} had any beneficial effects on accuracy (as opposed to robustness to adversarial attack). Second, while~\citet{safarani2021towards} reported improved robustness to image corruptions like adding noise and manipulating contrast, and they performed their analysis on CNNs trained to categorize ImageNet images, Safarani et al. did not investigate the impact of their ND regularizer on robustness to adversarial attack, and they did not find that the ND regularizer led to any noticeable improvements in accuracy. Finally, while ~\citet{Federer2020} showed improvements in accuracy and robustness to label corruption on the CIFAR-100 image recognition task, those performance gains were quite modest and they did not explore adversarial attack. Overall, the previous work leaves it unclear whether regularizers that force CNNs to mimic the brain's visual representations can achieve significant gains in accuracy, and in adversarial robustness, on categorization tasks involving complex natural images.

To address this open question, we build on the approach of ~\citet{Federer2020}, by training a CNN for object recognition while using a regularizer that forces the CNN's intermediate layers to form more brain-like image representations. In place of their representation similarity regularizer, which has serious limitations (discussed below), we developed a new regularizer that uses Deep Canonical Correlations Analysis (Deep CCA, DCCA~\citep{AndrewDCCA}) to quantify---and optimize---the similarity between the CNN's image representation and that of the brain. We find that CNNs trained on the CIFAR-100 task using our novel DCCA ND regularizer achieve larger gains in accuracy than those reported by~\citet{Federer2020}, and that our models are more robust to adversarial attacks than CNNs without the ND regularizer. Thus, our novel DCCA ND regularizer enabled us to overcome the limitations of the previous studies, and to advance the rapidly-growing field of ND regularizers in machine learning.

\section{Methods}
Our work builds on previous work~\citep{Federer2020} which regularized CNN training by pushing the CNN's representational similarity matrix (RSM)~\citep{kriegeskorte2008representational} to more closely resemble the RSM obtained from neural recordings of monkeys viewing images. The brain-RSM was comprised of the pairwise cosine similarity for vectors of neural firing rates from the monkey's neural recordings during image viewing; the CNN-RSM was comprised of the pairwise cosine similarity for the CNN's hidden representations for the same images. They included a term in the loss function that penalized the elements of the CNN-RSM if they were very different from the brain-RSM. While this approach can increase the similarity between the CNN and the neural data, it has several serious limitations that we enumerate here, and address with our new Deep CCA approach.

First, the CNN-RSM was computed using all units in a given CNN layer, while the brain-RSM was assembled from all  $\mathcal{O}(100)$ of the monkey's neurons \emph{observed} in the experiment. This means that the CNN's whole representation was being forced to resemble a tiny subset of the monkey neurons. Compounding this issue, the retinotopic organization of the visual cortex and the Utah array recording method used by \citet{coen2015flexible} means that this subset of neurons was sensitive to only part of the visual field.  The RSM method has no way to down-weight particular parts of the hidden representation if they do not represent the same information as the recorded neurons.

Second, the RSM matching approach, based on cosine similarity, cannot handle highly nonlinear relationships between the monkey brain's visual representation and that of the CNN.

Third, the neural recordings can contain noise, and so some of the recorded neurons may be unreliable. The cosine similarity metric used by~\citet{Federer2020} does not have the ability to differentially weight the recorded neurons, and so noisy neurons could drown out usable signals.

Our Deep CCA approach overcomes these three challenges, and is illustrated in Figure~\ref{fig:overview}.  We created a new architecture which combines a CNN with DCCA in order to incorporate information from the brain's responses to images. In the following section we first review Canonical Correlations Analysis (CCA) and then describe the deep neural network generalization---Deep CCA (DCCA)---that we use as a ND regularizer for training CNNs.

\subsection{Canonical Correlation Analysis (CCA)} 
Canonical Correlation Analysis (CCA) \citep{KantiBook} is a statistical method used to capture the relationship between two different views of the same data in a condensed form.  

In our scenario, these two views are the representation of an image created by a CNN (CNN-View) and the neural recording of the brain's response to that same image (Brain-View).  See Figure~\ref{fig:overview}. 

CCA finds coefficient vectors $\mathbf{a}_1$ and $\mathbf{b}_1$ that maximize the correlation between the two different views of the data, $\mathbf{x}$, $\mathbf{y}$ :
\begin{align}
\label{eq:cov}
\rho = \operatorname*{cov}(\mathbf{a}_1^T \mathbf{x},\mathbf{b}_1^T \mathbf{y})= \frac{\mathbf{a}_1^T \Sigma_{xy} \mathbf{b}_1}{\sqrt{\mathbf{a}_1^T \Sigma_{xx} \mathbf{a}_1} \sqrt{\mathbf{b}_1^T \Sigma_{yy} \mathbf{b}_1}}
\end{align}
Where $\Sigma_{xy} = \operatorname*{cov}(\mathbf{x},\mathbf{y})$ is the covariance matrix of random vectors {\bf{x}} and{ \bf{y}}, and 
$(\mathbf{a}_1^T \mathbf{x},\mathbf{b}_1^T \mathbf{y})$ is called the first canonical pair. 
The subsequent pairs of coefficient vectors are found by identifying vectors that optimize Eq.~\ref{eq:cov} and are orthonormal to all previous pairs of coefficient vectors. The total number of canonical pairs found is controlled with hyperparameter $C$.  The relationship between these two views, $\mathbf{x}$ and $\mathbf{y}$, is described by the set of $C$ \emph{canonical} correlations, indexed by $i$, between $(\mathbf{a}_i^T \mathbf{x},\mathbf{b}_i^T \mathbf{y})$.

While CCA can overcome the \emph{neuron subsets} problem (the need for differential weighting of neurons with different reliability, faced by the RSM method of~\citet{Federer2020} discussed above), it shares the RSM method's limitation of being insensitive to highly nonlinear relationships between the neural representation and that of the CNN. To overcome this challenge, we use a nonlinear generalization of CCA.

\subsection{Deep CCA (DCCA)}
Deep CCA~\citep{AndrewDCCA} is a deep neural network (DNN) version of CCA in which the two views of data ($\mathbf{x}$ and $\mathbf{y}$ from Eq. \ref{eq:cov}) are passed through separate and independent DNNs to yield nonlinearly-transformed representations of those views. The final DNN representations of the data views are then used as input to the typical CCA optimization procedure to minimize the negated sum of the $C$  canonical pairs' correlations.  This loss is then used to backpropagate and update the weights of each DNN.  We have loss function:
\begin{align}
\mathcal{L_\mathrm{DCCA}} = 
 -  \operatorname*{CCA}(f_{\textbf{x}}(\textbf{x}, \Theta_\textbf{x}), f_{\textbf{y}}(\textbf{y}, \Theta_\textbf{y}))\label{eq:DCCA}
\end{align}
where $\Theta_\textbf{x}, \Theta_\textbf{y}$ are the weights and biases from the respective DNN sub-models $f_\textbf{x}(\cdot)$ and $f_\textbf{y}(\cdot)$ of the DCCA branch, and CCA is the sum over all canonical pairs using Eq. \ref{eq:cov}.

The two DCCA sub-models have hyperparameters (activation functions, dropout rates, and learning rates), which are tuned and may differ between the sub-models. In Figure~\ref{fig:overview} the two DCCA sub-models take input from the CNN-View and Brain-View.  Compressing the data and including nonlinear activation functions allows us to find more complex transformations of the data. 

\subsection{Training CNNs with a Deep CCA Neural Data Regularizer}\label{s.trainDCCA}

\begin{figure}
    \centering
    \includegraphics[width=14cm]{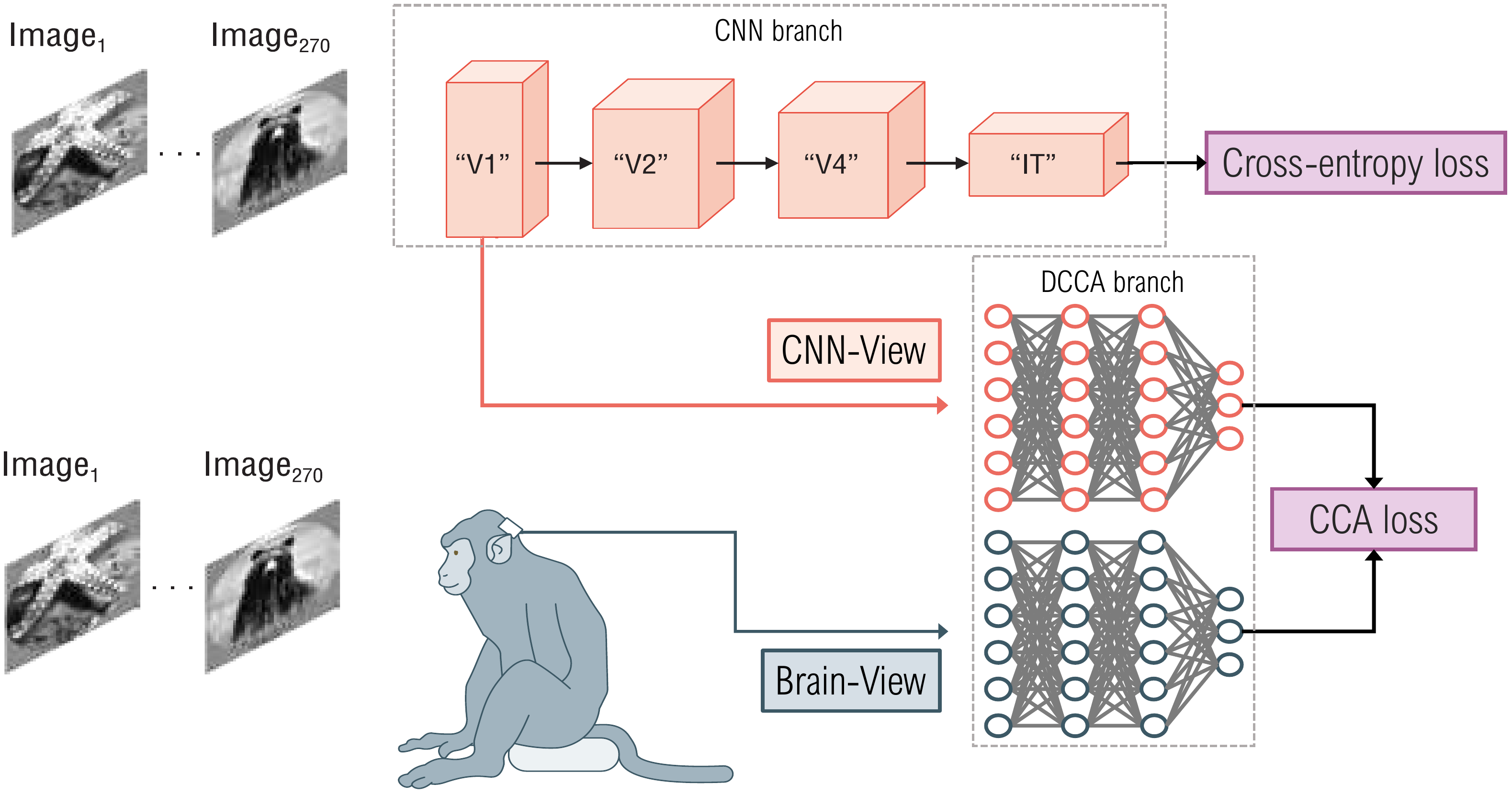}
    \caption{An overview of the ND regularizer architecture.  Recordings of brain activity (here, direct neural recordings from a monkey) during stimulus image viewing are passed as the Brain-View of the DCCA branch.  The same stimulus image is passed to the CNN branch (CORnetZ), and the activations from the V1 layer are passed as the CNN-View to the DCCA branch.  CCA loss is computed with the two final representations from the two DCCA sub-models.  Cross-entropy loss is computed using CIFAR-100 images.  Weights used to compute V1 representations are impacted by both loss functions.  Figure adapted from~\citet{Federer2020}.}
    \label{fig:overview}
\end{figure}

We used Keras to create a CORnetZ model (originally developed by \citet{kubilius2018cornet}) linked to a DCCA model. CORnetZ has a structure that maps onto the hierarchy of the visual system. We used the hidden representations in CORnetZ's V1 layer as the CNN-view for the DCCA branch depicted in Figure~\ref{fig:overview}. We chose the V1 layer because the Brain-view is obtained from neural recordings of the V1 area in the monkey primary visual pathway. These recordings are from \citet{coen2015flexible} and are described in more detail in Section~\ref{sec:data}.

For a given brain-recording training sample, the image that was shown to the monkey in the neuroscience experiment is fed through CORnetZ's V1 layer, and the activations are fed into the CNN-View of the DCCA model.  At the same time, the neural response to that same image becomes the Brain-View for the DCCA model.  Batches of 50 image/neural-recording pairs are passed through the two DCCA sub-models, and the final representations of the two data stream batches are used in the CCA loss function, which is completely differentiable.  Backpropagation proceeds through both sub-models, including the portions of CORnetZ that are used to create the CNN-View representations.

The CNN uses cross-entropy (CE) as its loss function, which we add to the DCCA loss to create a joint loss function $\mathcal{L}$ .  We control how much each loss contributes to the joint loss using hyperparameter $\lambda$. Because we normalize the DCCA cost, $\lambda$ can be described as the proportion the DCCA cost is of the total cost. For example, $\lambda$ = 1 would mean that the total cost is just the DCCA cost, alternatively $\lambda$ = 0 means the total cost is just the CE cost. 
\begin{align}
 \mathcal{L}_\mathrm{CE} &= \sum_i \hat{p_i} \log(p_i)
 \end{align}
 \begin{align}
 \mathcal{L}_\mathrm{DCCA} &=  - \operatorname*{cov}\big (f_{\textbf{x}}(\textbf{x}), f_{\textbf{y}}(\textbf{y}) \big ) / C
 \end{align}
 \begin{align}
 \mathcal{L} &= \lambda\ \mathcal{L}_\mathrm{DCCA} + (1-\lambda)\ \mathcal{L}_\mathrm{CE} 
\end{align}
where $p_i$ is the indicator variable for the true class of the image and $\hat{p}_i$ is the probability the model assigns to the true class.
We scale $\mathcal{L}_\mathrm{DCCA}$ by $C$, the number of canonical pairs, so that it has the same magnitude as $\mathcal{L}_\mathrm{CE}$.

The stimuli images used by \citet{coen2015flexible} do not have labels associated with them, so we cannot use them to compute $\mathcal{L}_\mathrm{CE}$.  Instead, $\mathcal{L}_\mathrm{DCCA}$ is computed using the stimuli images from the monkey experiment (along with the neural recordings) and $\mathcal{L}_\mathrm{CE}$ is computed with images from CIFAR-100. 

Because the dataset sizes differ, for each epoch of training on CIFAR-100, we cycle through the neural data 20 times. In our figures the epochs count on the horizontal axes refers to the epochs of training on the CIFAR-100 data.

\subsection{CORnetZ}\label{sec:Cornet}
The CORnet family of network architectures was created to represent the primate visual cortex by \citet{kubilius2018cornet}. The CORnet family uses simple layers that all refer to an area of the primate visual pathway and performs and behaves in a manner such that they match primate brain and behavioral data. 

In this work, we use the simplest member of this family, the CORnet-Zero, or CORnetZ. CORnetZ has 4 areas (V1, V2, V4, IT), each consisting of only a single convolution, followed by a ReLU nonlinearity and max pooling. These are followed by a single 100-way linear classification layer.

\subsubsection{Data}\label{sec:data}
We use the popular CIFAR-100 \citep{cifar10} dataset (no license specified) for our image classification task. This dataset has 600 images for each of its 100 classes resulting in a total of 60 000 images.  CIFAR-100 also further, evenly categorizes the images into 20 super-classes such that one super-class contains 5 of the sub-classes. We will later quantify the \emph{exact-class} accuracy and the \emph{super-class} accuracy of our trained networks: these quantities will be measured on the same trained networks. Exact-class accuracy is defined as the fraction of images for which the CNN returns the correct class, while super-class accuracy is the fraction of images for which the CNN returns the correct super-class.

For the DCCA ND regularizer, we use publicly available and previously published neural data from \citet{coen2015flexible}. This data was collected in the Laboratory of Adam Kohn at the Albert Einstein College of Medicine and downloaded from the CRCNS website (no licence specified). These neural recordings were obtained by inserting Utah arrays into the primary visual cortex (V1) of anaesthetized monkeys: we used all the available data, which was from 3 individual monkeys. The Utah arrays record neural spiking activity from $\mathcal{O}$(100) neurons at a time. In one session this data was recorded while the monkeys were shown a series of natural images (270 images) and static gratings. Each image was presented for 100\,ms and repeated 20 times. 

This recording procedure was repeated in 10 different experimental sessions. Though multiple sessions were recorded for each monkey, the $\mathcal{O}$(100) neurons recorded in each session are not always the same. For sessions with different monkeys, the neurons are definitely not analogous.  In general, we assumed each session recorded the responses of different groups of neurons to the same set of 956 images (including both the natural images and static gratings).

We preprocessed this data by averaging the neural spike counts for each image presentation across the 20 repeats. Next, we made one pseudo-population out of the neural recordings on the 10 individual sessions. This pseudo-population consists of the top 80 principal components of the neural responses from each session, concatenated together. Thus, the final preprocessed neural activity data had dimensions of 956 images $\times$ (80 $\times$ 10) principal components.

\section{Experiments}\label{sec:Experiments}
To determine the effects of the ND regularizer on CNN performance, we trained CNNs with different regularization strengths $\lambda$.  For the CNN training, we used the same settings as \citet{Federer2020}: learning rate 0.01, CIFAR-100 batch size of 128, and dropout rate of 0.5.  

Each of the DCCA sub-models consists of 3 dense layers of width 1024, followed by a 0.0001 dropout layer, and finally a dense layer with a width of 10. We then compute CCA using these two final outputs from each DCCA sub-model.   

The DCCA-branch of the network required independent hyperparameter tuning, which we performed using the DCCA-branch alone (no CNN-branch). We optimized the DCCA loss function, which is the negated sum of the correlations of the $C$ canonical pairs. We found the optimal settings to yield high DCCA correlation were: batch size 50, ReLU activation, 10 canonical pairs, along with 0.00001 L2 weight decay and a Random Normal kernel intitializer with standard deviation of 0.01. We used these hyperparameters when training the combined CNN + DCCA model. We trained the full CNN + DCCA models for 100 epochs and performed 5 randomly-seeded repetitions for each different regularization strength $\lambda$.  We experimented with pretraining the DCCA before incorporating it into the CNN, and found only a modest improvement. Because pretraining the DCCA adds computational overhead, and the effect on accuracy is small at best, we did not pursue this approach.

As a control, to determine if any of the performance gains obtained with the ND regularizer depended on the specific response of the brain to an image, we repeated our CNN regularization experiments with randomized neural data in which the image labels associated with each measured neural response were randomly permuted. The shuffled dataset matched each neuron's distribution of firing rates but removed any of the neuron's sensitivity to specific image features. Thus, the difference in performance obtained with the un-permuted neural data and with this permuted data, reveal the extent to which the brain's image representations (as opposed to its other statistical properties) are a useful regularizer for CNNs.

\subsection{The Effect of the Neural Data Regularizer on Accuracy}

\begin{figure}
     \centering
     \begin{subfigure}[b]{0.49\textwidth}
         \centering
         \includegraphics[width=\textwidth]{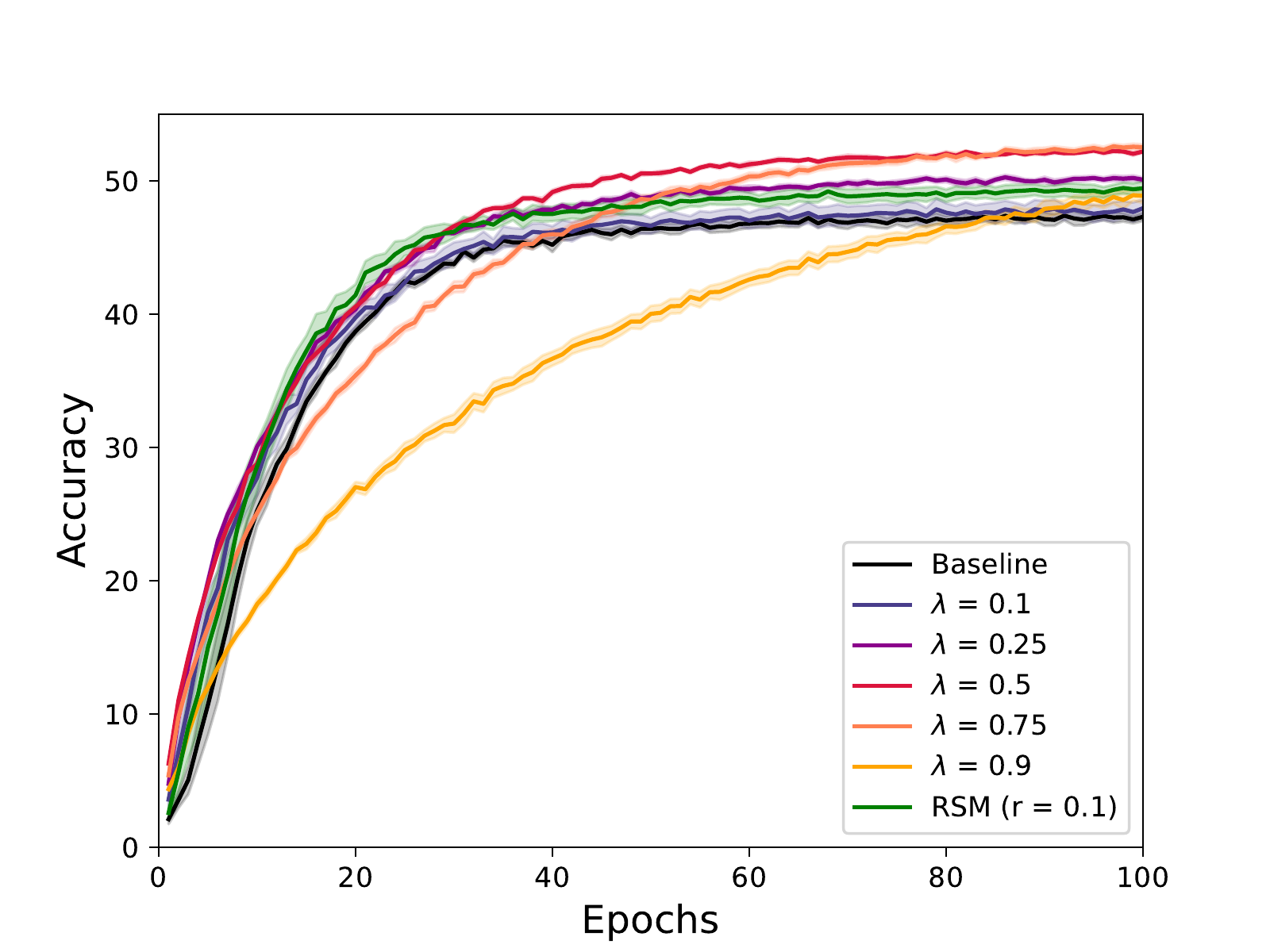}
         \caption{Validation Accuracy on CIFAR-100 over epochs for different regularization strengths $\lambda$.}
         \label{fig:accuracy}
     \end{subfigure}
     \hfill
     \begin{subfigure}[b]{0.48\textwidth}
         \centering
         \includegraphics[width=\textwidth]{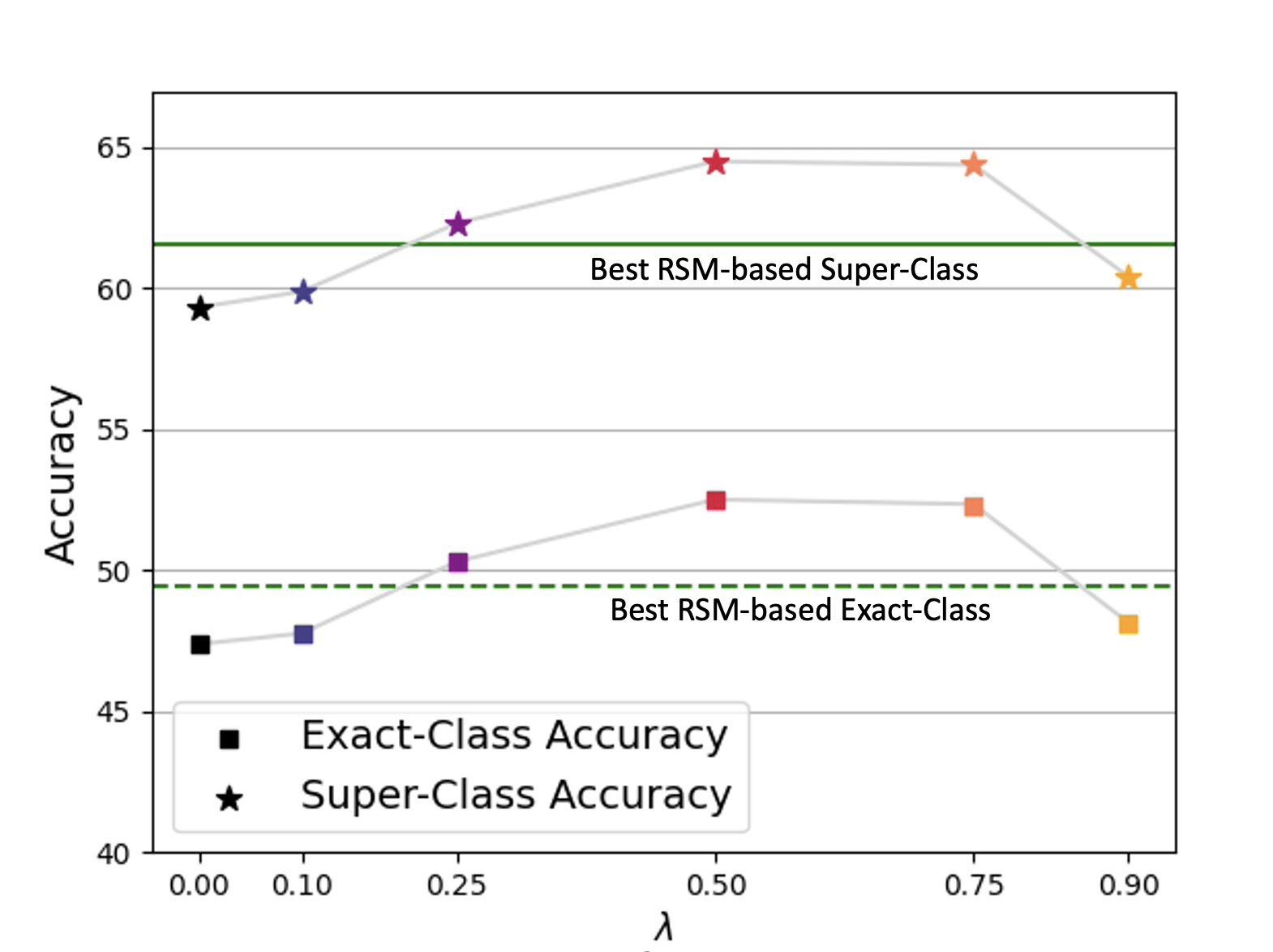}
         \caption{Exact-class and super-class accuracy after training for 100 epochs for various $\lambda$ values}
         \label{fig:superclass}
     \end{subfigure}
     \caption{Measuring Accuracy on CIFAR-100 for different regularization strengths.  {\bf (a)} Validation accuracy during training.  {\bf (b)} Exact-class and super-class accuracy after training for 100 epochs for various $\lambda$ values, $\lambda=0$ is baseline (Colors match legend in {\bf (a)}). Error bars did not extend past the markers, and so are not visible.
    Exact-class accuracy rises for increasing $\lambda$ until it starts to drop at $\lambda$ = 0.75. Super-class accuracy is higher than the exact-class accuracy and shows a maximum at $\lambda=0.50$. RSM performance mentioned here are from \citet{Federer2020} with their best performing r value (r = 0.1).}
\end{figure}

We tested the impact of our regularizer on image classification accuracy using CIFAR-100 and CORnetZ (Fig.~\ref{fig:accuracy}).  
We found the optimal $\lambda$ to be $0.5$ at 100 epochs, with accuracy $52.3\%$, though the performance of $\lambda=0.75$ is very close.  This outperforms both the baseline ($\lambda=0$, $47.4\%$) and previous ND regularizers, which reported maximum accuracy of $49.5\%$~\citep{Federer2020}.  For $\lambda \geq 0.5$, the performance of the regularized model dominates over the baseline even very early in training.  For high values of $\lambda$ the accuracy drops below the baseline, but overtakes it later in training.
Note also that the baseline accuracy appears to plateau at around 40 epochs, but the regularized models continue to improve by $\sim 5\%$ even after that point. We also tried only including the regularizer ($\mathcal{L}_\mathrm{DCCA}$) for a subset of all epochs, at the start, middle, and end.  We found that using DCCA for the entirety of the training provided the best results. 

To test our ND regularizer on a better performing model than CORNetZ, we used ResNet-50 \citet{resnet}. We found that smaller $\lambda$ ($= 0.1$) values achieved higher accuracy. We find that the regularizer can be less beneficial or even detrimental to a model that already builds precise representations when $\lambda$ values are too high. For more on these experiments see Supplementary Materials (Section \ref{sec:si_resnet}). 

To what attributes of the neural data can we attribute this increase in accuracy? To answer this question we generated datasets with varying similarities of the neural data and re-ran our analysis. First, we shuffled the labels that mapped the recordings to the images. This preserves the neural recording's latent structures and features (such as the distribution) in response to a visual processing task while disconnecting the recording from the specific image that the subjects viewed. Next, we generated data using a normal distribution with the same mean and standard deviation as the neural data (V1 statistics). Lastly, we generated a random dataset following a standard normal distribution.

We found that, compared to the neural data, the best $\lambda$ for the shuffled dataset produced an ND regularizer that was almost as effective in terms of accuracy (see Fig.~\ref{fig:shuffled}). A similar performance was found for the other generated datasets. However, note that the best performances are in order from most brain-like to least: neural data, shuffled data, V1 statistics data, and lastly standard normal data.  This is consistent with the findings of \citet{Federer2020}, who found that it was the high-level statistics of the neural data that seemed to drive the effect, rather than the individual firing rates for each image. This finding deserves further exploration, which we leave for future work. However, as we will see later, the network trained on un-shuffled neural data is more robust to adversarial attack. This suggests that the brain's detailed image representations, and not just those representations' statistical properties, contribute to robustness of the CNN performance. Here we will do further analysis with the shuffled data, for more analysis on the other generated datasets see the Supplementary Materials (Section \ref{sec:si_datasets}.

\subsection{The Effect of the Neural Data Regularizer on Super-Class Accuracy}
We have determined that the accuracy of the model is improved with ND regularization, but the model still makes mistakes. Are the mistakes made by the ND regularized model more reasonable?  To test this, we measured the super-class accuracy of the trained models.  Here, we re-calculate accuracy at the super-class level, rather than the exact-class level.  For example, if an image of a mouse is misclassified as a hamster, the super-class accuracy would count this as a correct classification because they share the same CIFAR-100 super-class (small mammal).  But, if a mouse is misclassified as a possum, this would be a misclassification for both exact-class and super-class accuracy, as the possum is considered a medium-sized mammal in CIFAR-100. 

Figure~\ref{fig:superclass} shows the final exact-class and super-class accuracy of models trained with a variety of $\lambda$ values. The super-class accuracy of the baseline model is 59\% and the best performing ND regularized model is $64.5\%$.  This is consistent with the findings of \citet{Federer2020}, who found that ND regularization had a positive impact on both exact-class and super-class accuracy.

\subsection{The Effect of the Neural Data Regularizer on Adversarial Robustness}
Previous work has shown that small perturbations in the input image can lead to misclassifications by the CNN, even when the perturbations are nearly imperceptible to a human~
\citep{goodfellow2014explaining}.  We wondered if our ND regularizer might provide some protection against such attacks.

To test our networks' robustness against adversarial attacks, we used the fast gradient sign method (FGSM, \citep{goodfellow2014explaining}). The FGSM calculates the derivative of the loss function with respect to the input pixels, and takes the sign of these gradients to provide a change to the input image that maximizes the loss. This change to the image is multiplied by a strength factor and added to the image to provide an attack image.

\begin{figure}[htbp]
    \centering
    \includegraphics[width=8cm]{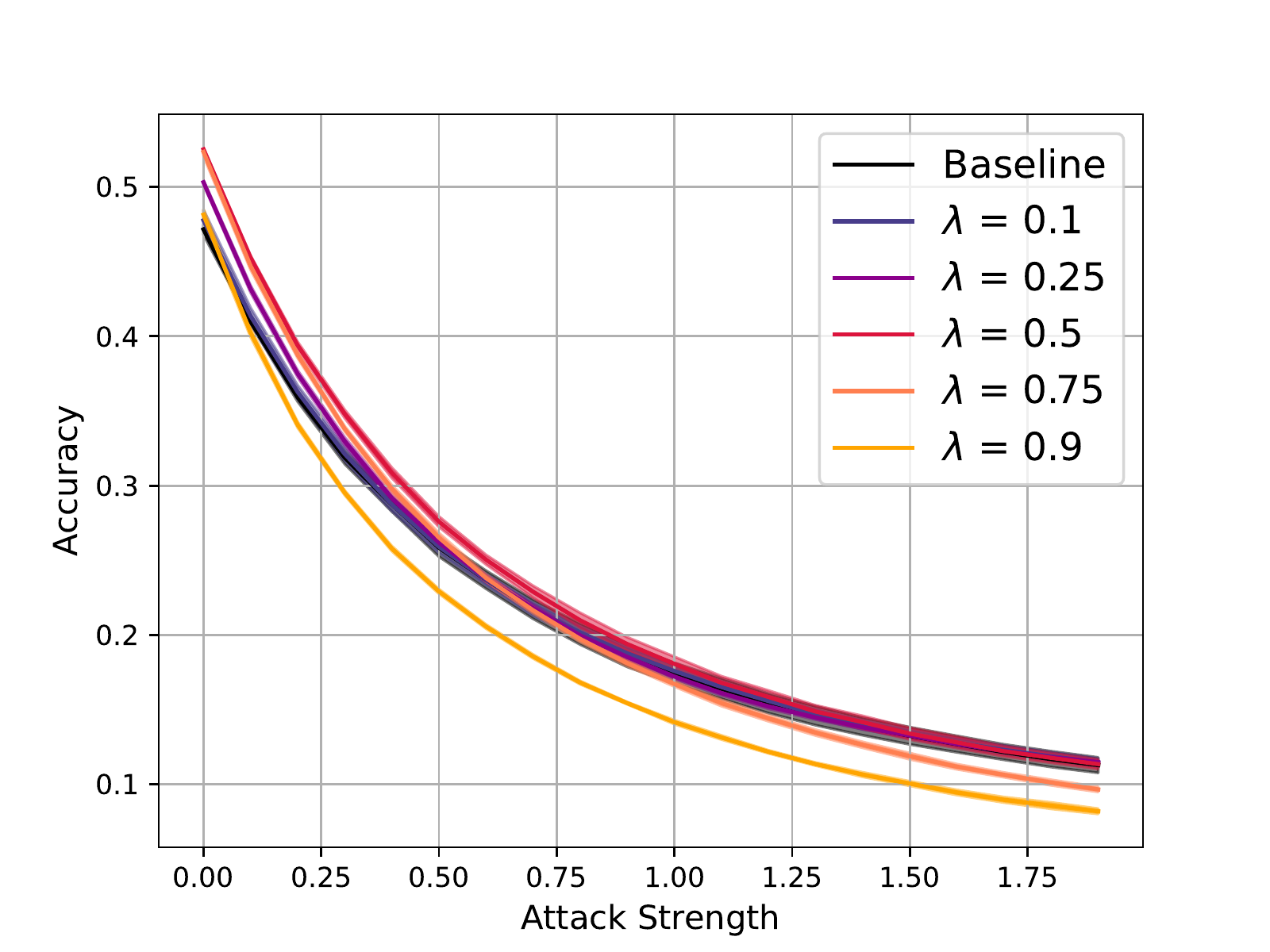}
    
    \caption{Robustness against adversarial attack for FGSM attacks for the baseline network (black) compared to the ND regularized networks for different values of $\lambda$. Robustness is higher than baseline for regularized networks when the attack strength is low to moderate. 
    }
    \label{fig:attacks_all}
\end{figure}

\begin{figure}[htbp]
    \centering
     \begin{subfigure}[b]{0.49\textwidth}
         \centering
         \includegraphics[width=\textwidth]{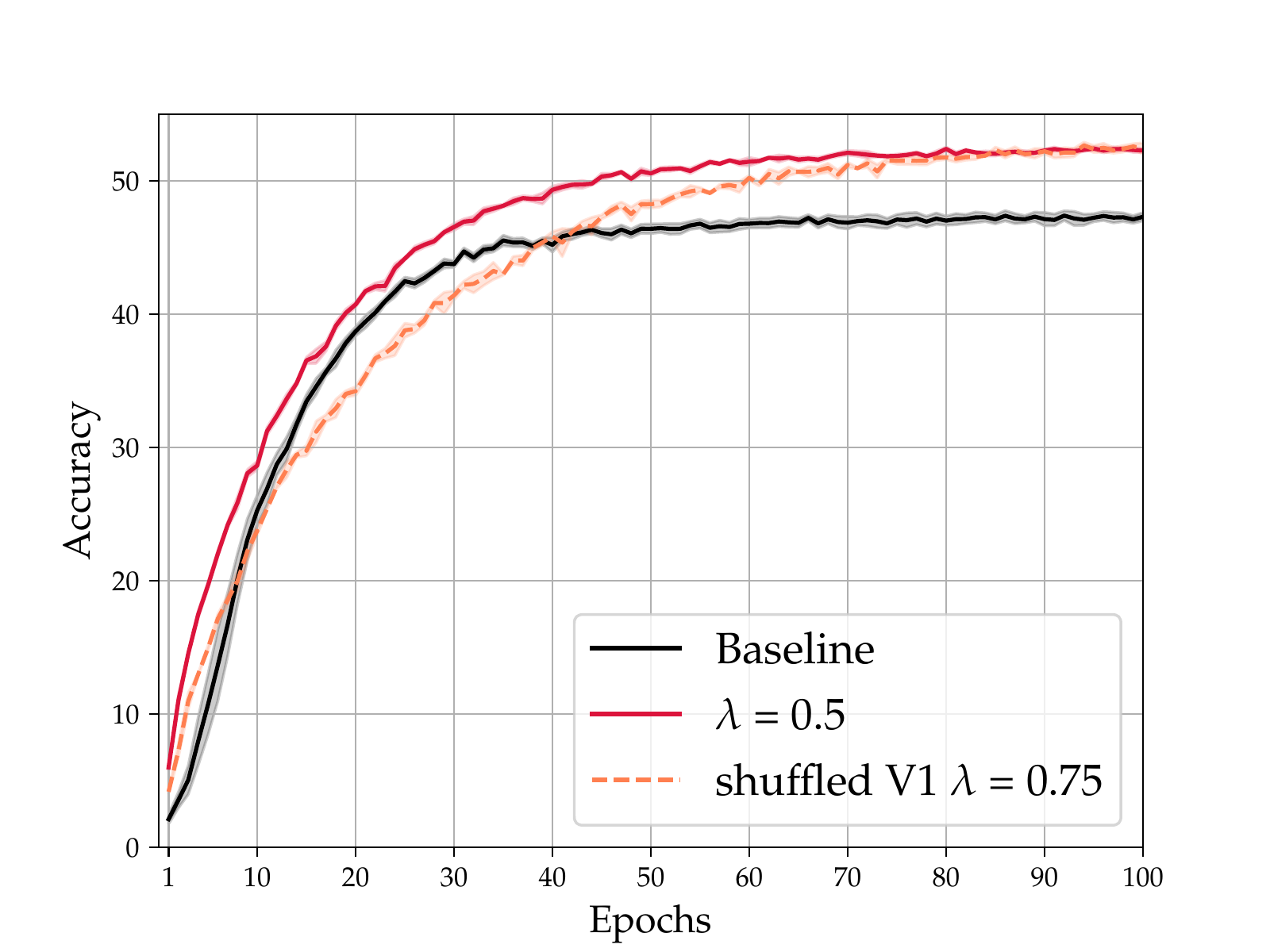}
         \caption{}
          \label{fig:shuffled}
     \end{subfigure}
     \hfill
     \begin{subfigure}[b]{0.49\textwidth}
         \centering
         \includegraphics[width=\textwidth]{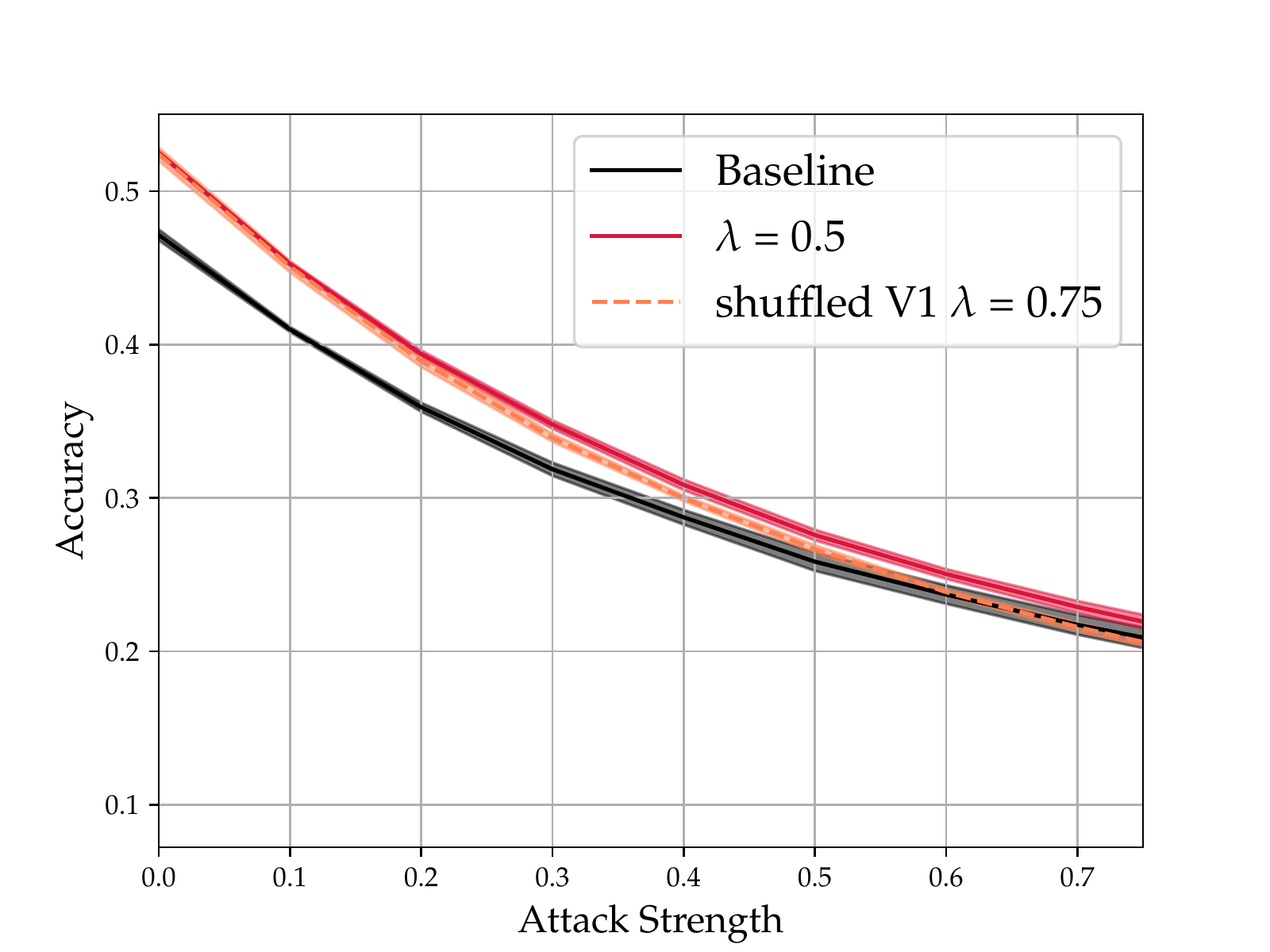}
         \caption{}
          \label{fig:adverse_shuffled}
     \end{subfigure}
    \caption{{\bf(a)} A comparison of the best-performing ND regularized networks using shuffled and un-shuffled neural data.  For un-shuffled neural data, $\lambda=0.5$, for shuffled neural data, $\lambda=0.75$. At 100 epochs, the accuracies are within $0.4\%$ of each other. {\bf (b)} The robustness of the best-performing ND regularized networks (from panel {\bf a}) to adversarial attacks.  Though the networks performed similarly in terms of accuracy, the un-shuffled ND regularized network is more robust to adversarial attack. For performance on stronger attacks, see Supplementary Material (Fig.\ \ref{fig:si_adverse_shuffle}). Standard deviation are shown as shaded lines around the mean lines.} 
\end{figure}

Our ND regularized networks show a clear robustness for moderate strength attacks (Fig. \ref{fig:attacks_all}). For stronger attacks the difference vanishes, and the images are no longer recognizable for any network. Figure~\ref{fig:adverse_shuffled} shows that, though the accuracy using shuffled and un-shuffled neural data was similar (Fig.~\ref{fig:shuffled}), using the non-shuffled data in the ND regularizer produces a network that is more robust to adversarial attack.  Thus, the brain's detailed image representations (not just those representations' statistical properties) provide some protection from adversarial attack.

\section{Limitations}\label{sec:Limitations}
There are limitations to our work. Firstly, the neural recordings used here were collected using anesthetized monkeys viewing images of things they may not have encountered or experienced (e.g. images of European landmarks).  The ND regularizer may be more effective if the neural data came from an awake and motivated animal or a human who is familiar with the objects in the stimuli images.  In addition, the number of neurons captured in each monkey is small ($\mathcal{O}$(100)), and so doesn't represent the full range of activity in V1.
 
We found that a regularizer using the shuffled neural data worked as well as one using un-shuffled neural data.  Though this is consistent with previous findings \citep{Federer2020}, it does leave some questions about the origin of the effect unanswered.

\section{Conclusion}
We have outlined a new method for incorporating neural recordings as a source of regularization for CNNs.  We have shown that our ND regularizer outperforms both a baseline model, and previously published ND regularizing techniques.  We also showed that regularized networks produced better super-class accuracy and were more robust to adversarial attack.  In sum, our results provide evidence for a promising new area of research, incorporating neural data to improve the representations of neural network models.

In the future we would like to explore neural data captured from humans engaged in an image-related task (e.g.\ a one back task).  We would also like to explore the effect of our regularizer on other networks, and hope to show similar improvements in larger architectures.   

\section{Ethical Concerns and Negative Societal Impacts}\label{sec:ethics}
All neural recordings are from a previously released public dataset~\citep{coen2015flexible}.  The recordings were collected with the oversight of the institutional review boards at the institute where the experiments were performed. Researchers followed the guidelines in the United States Public Health Service Guide for the Care and Use of Laboratory Animals.

Here we trained on the fairly small CIFAR-100 dataset with a smaller network, so the immediate ethical considerations are  minimal.  However, our hope is that these techniques can be extended to improve the accuracy of a wide variety of networks. At that point, the ethical implications would become more significant.

Because our method incorporates neural data into CNNs, this might allow some undesirable signatures of meaning representations to be transferred to a model.  For example, if the neural data was collected from a person who held (conscious or unconscious) biases against particular racial groups, it is possible that such biases could be transferred into the CNN. At this point, we have worked only with neural recordings from monkeys, but future work may incorporate human neural recordings. At that point, it will be important to monitor research on detecting and mitigating bias in models, and incorporate such approaches into our training pipeline. 

\section{Software}
\label{sec:software}
We ran our experiments using the Keras API \citep{chollet2015keras} with Tensorflow backend \citep{tensorflow2015-whitepaper} in Python \citep{RossumPython}. We utilized the Python packages Numpy and Pandas to analyze our findings, and the packages Matplotlib \citep{Hunter:2007} and Pylustrator \citep{Gerum2019e} to create visual reportings of our results. We based our DCCA implementation off of \url{https://github.com/davidfsemedo/DCCA-keras-tensorflow/blob/master/} (MIT License)  Experiments were run on a Shared Cluster including NVIDIA P100 and V100 GPUs.

\bibliographystyle{apalike}
\bibliography{references}

\appendix

\newpage
\renewcommand\thefigure{S\arabic{figure}}
\renewcommand\thesection{S\arabic{section}}
\renewcommand\thesubsection{S\arabic{section}.\arabic{subsection}}
\setcounter{figure}{0} 

\section{ResNet-50 experiments}
\label{sec:si_resnet}
To determine whether our ND regularizer works on a more powerful model than CORnetZ we used ResNet-50 as our CNN. We use the same hyperparmeter tuning for the DCCA model as in Section 4. We used the architecture described in \citet{resnet} along with an image size of 32, batch normalization momentum of 0.9, and an orthogonal kernel initializer.  Image augmentation of random horizontal flips and rotations was also used in training.  The ResNet-50 experiments were ran for 200 epochs with a learning rate schedule as follows: learning rate = 0.1 for epochs 1-60, learning rate = 0.02 for epochs 61-120, learning rate = 0.004 for epochs 121-180, and learning rate = 0.0008 for epochs 181-200.  

Smaller $\lambda$ values were experimented with as we suspected the ResNet-50 model was already achieving sophisticated and brain-like representations (\citet{yamins}) on its own. Three random seeds were run for each $\lambda = (0.01, 0.1, 0.25, 0.5)$. 

According to Brain Score (\citet{b_score}), a composite benchmark for comparing models to the brain, the last convolutional layer of the second block forms the representations most similar to the brain. For this reason we chose to connect the DCCA to this node.

We found that the highest accuracy was achieved from using the ND regularizer for the first 10 epochs only, similar to \citet{Federer2020}.  Note that using our configurations we were unable to achieve state of the art results with the ResNet-50 baseline \citet{resnet_state} (for example, transfer learning was not used for our model). 

\begin{figure}[htbp]
\centering
    \begin{subfigure}[b]{0.49\textwidth}
         \centering
         \includegraphics[width=\textwidth]{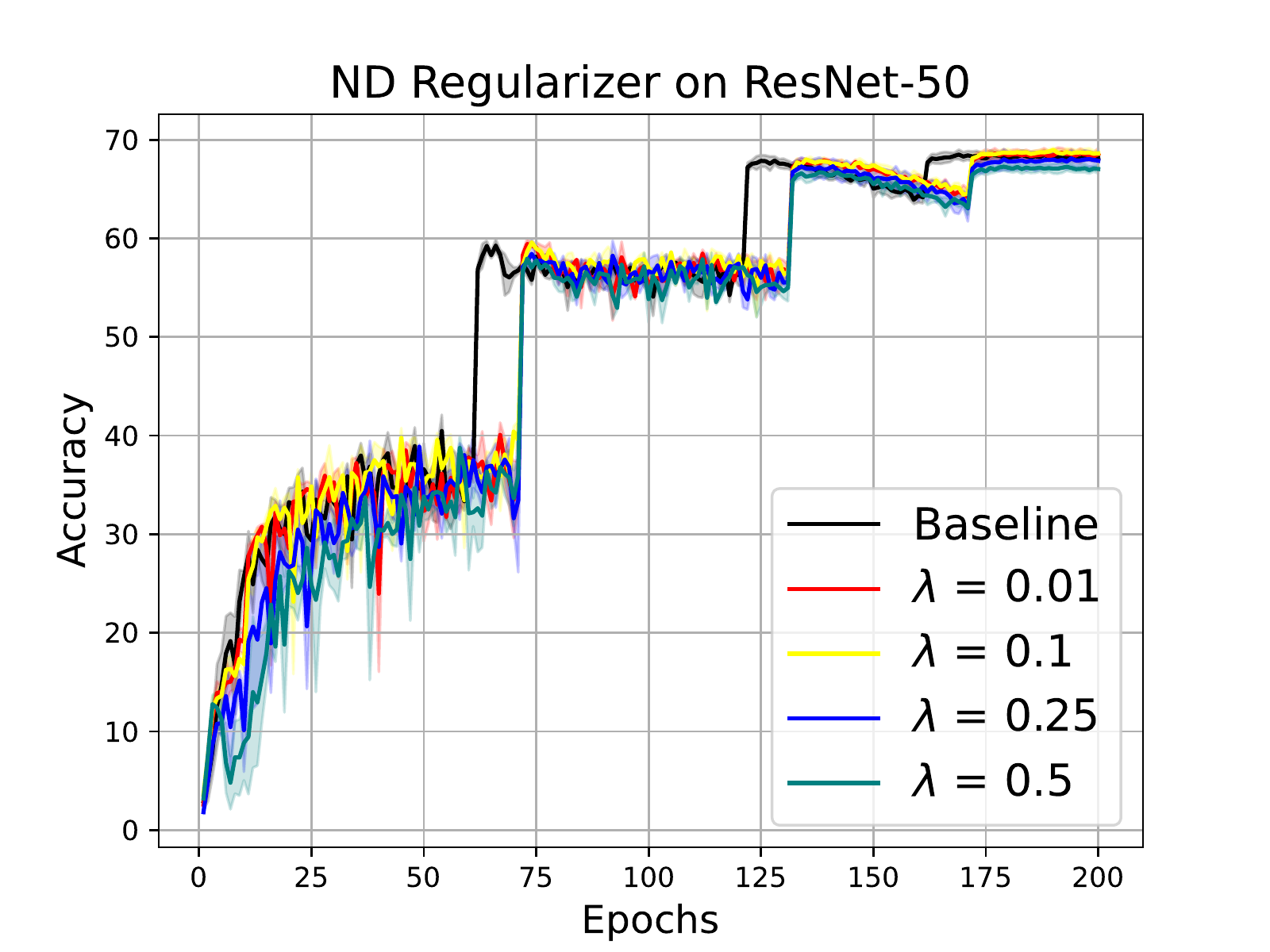}
         \caption{}
     \end{subfigure}
     \hfill
     \begin{subfigure}[b]{0.49\textwidth}
         \centering
         \includegraphics[width=\textwidth]{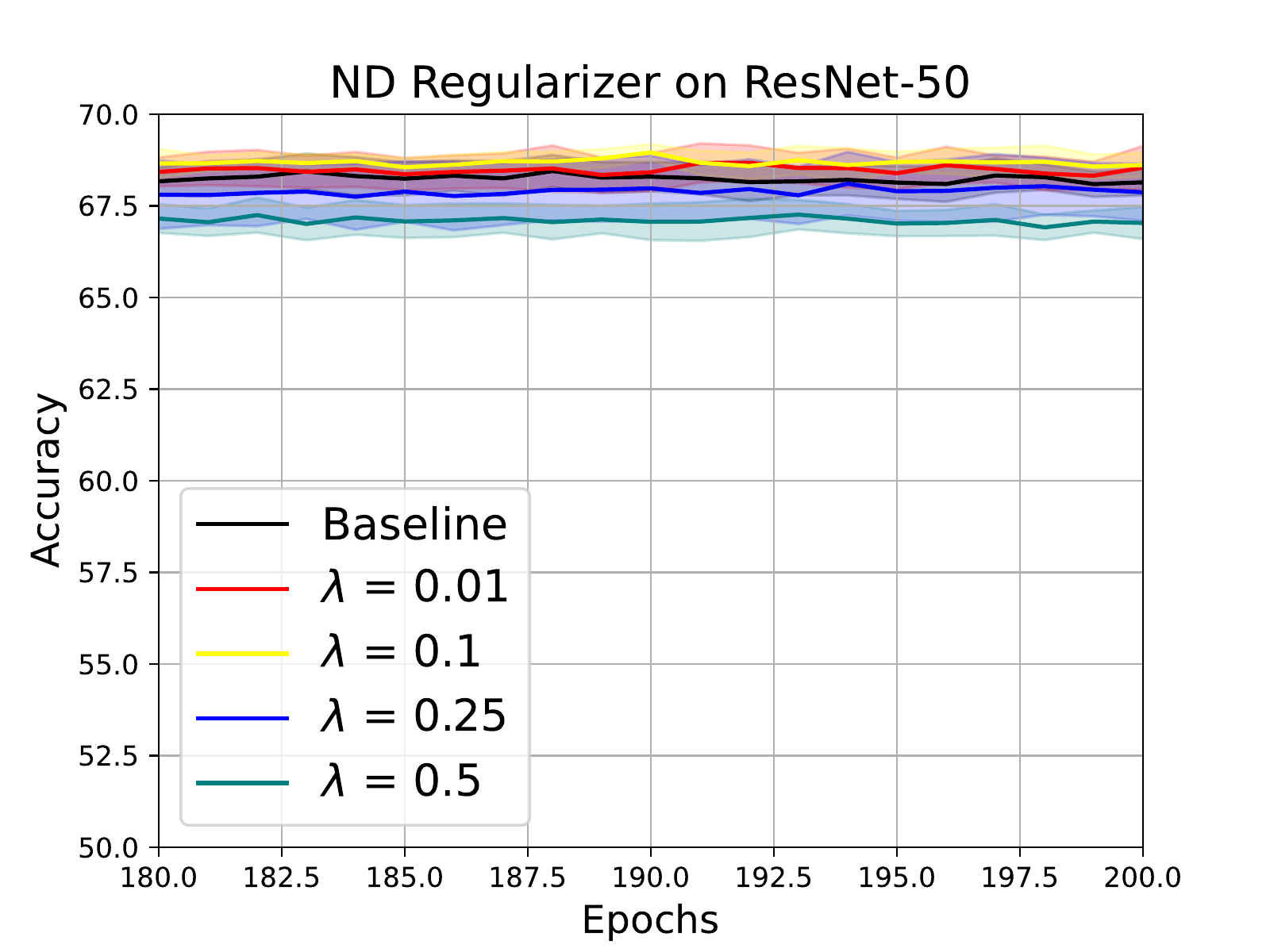}
         \caption{}
     \end{subfigure}    
    \caption{Accuracy experiment on ResNet-50 with CIFAR-100 data. Both subfigure \textbf{[a]} and \textbf{[b]} are of the same data with axes starting at different points. It can be seen in figure \textbf{[b]} that $\lambda = 0.01, 0.1$ consistently achieve higher accuracy than the unregularized model (68.14\%). $\lambda = 0.01$ has accuracy of 68.54\% on the 200th epoch, $\lambda = 0.1$ has accuracy of 68.61\% on the 200th epoch. The larger $\lambda$ values (0.25, 0.5) achieve lower accuracy than the unregularized model.}
    \label{fig:si_resnet}
\end{figure}
\clearpage

\section{Robustness to Adversarial Examples}
\begin{figure}[htbp]
    \centering
    \includegraphics[width=0.5\textwidth]{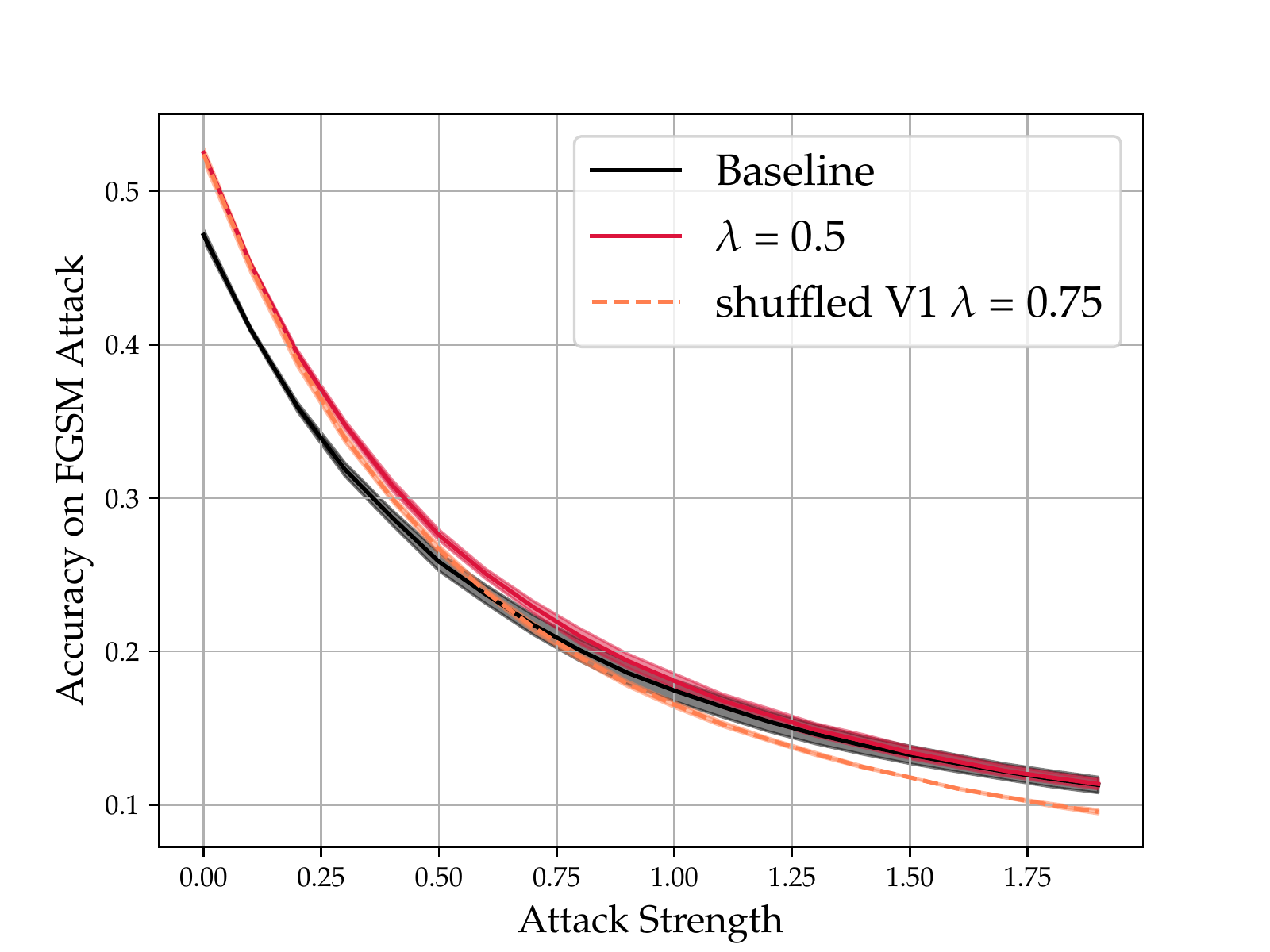}
    \caption{Robustness of the best-performing
ND regularized networks ($\lambda = 0.5$ un-shuffled and $\lambda = 0.75$ shuffled) to adversarial attacks. We see the regularized networks are more robust than the un-regularized network for attack strengths up to 0.75. For higher attacks, the un-shuffled regularized network performs slightly better than the un-regularized network, but the shuffled regularized network's performance drops below the un-regularized network.}
    \label{fig:si_adverse_shuffle}
\end{figure}

\clearpage
\section{Generated Datasets}
\label{sec:si_datasets}

\subsection{Accuracy}
\begin{figure}[hbtp]
    \centering
    \begin{subfigure}[b]{0.49\textwidth}
    \includegraphics[width=\textwidth]{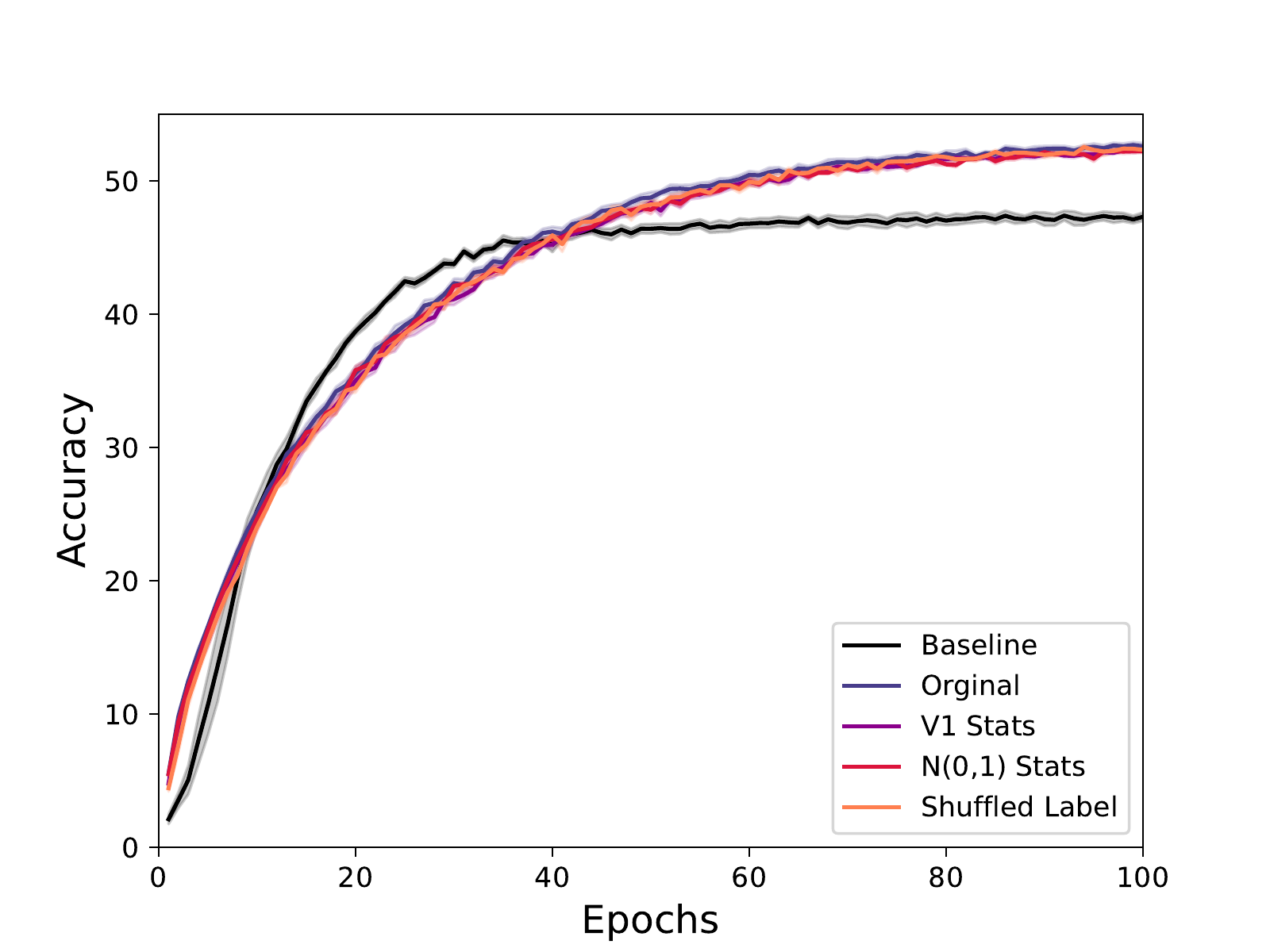}
    \caption{}
    \end{subfigure}
    \begin{subfigure}[b]{0.49\textwidth}
    \includegraphics[width=\textwidth]{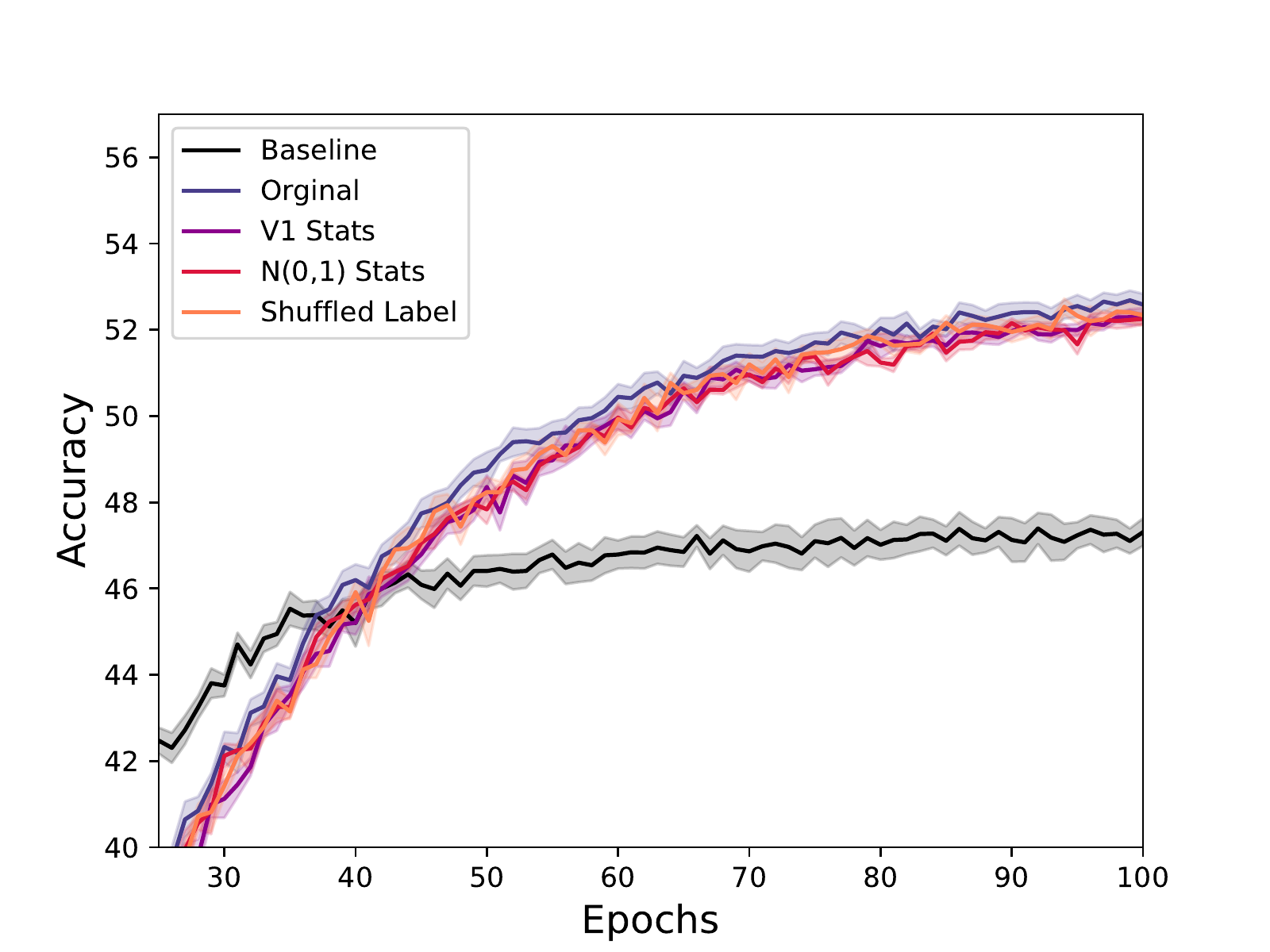}
    \caption{}
    \end{subfigure}
    \caption{Accuracy experiment on the different datasets. The best performing $\lambda$ for all datasets was $\lambda = 0.75$. We see with the ND regularizer the original and the generated the  datasets have the same trend. Closer inspection shows that the original dataset does slightly better than the others in terms of accuracy.}
    \label{fig:si_generated_acc}
\end{figure}

\subsection{Super-Class Accuracy}

\begin{figure}[hbtp]
    \centering
    \includegraphics[width=0.49\textwidth]{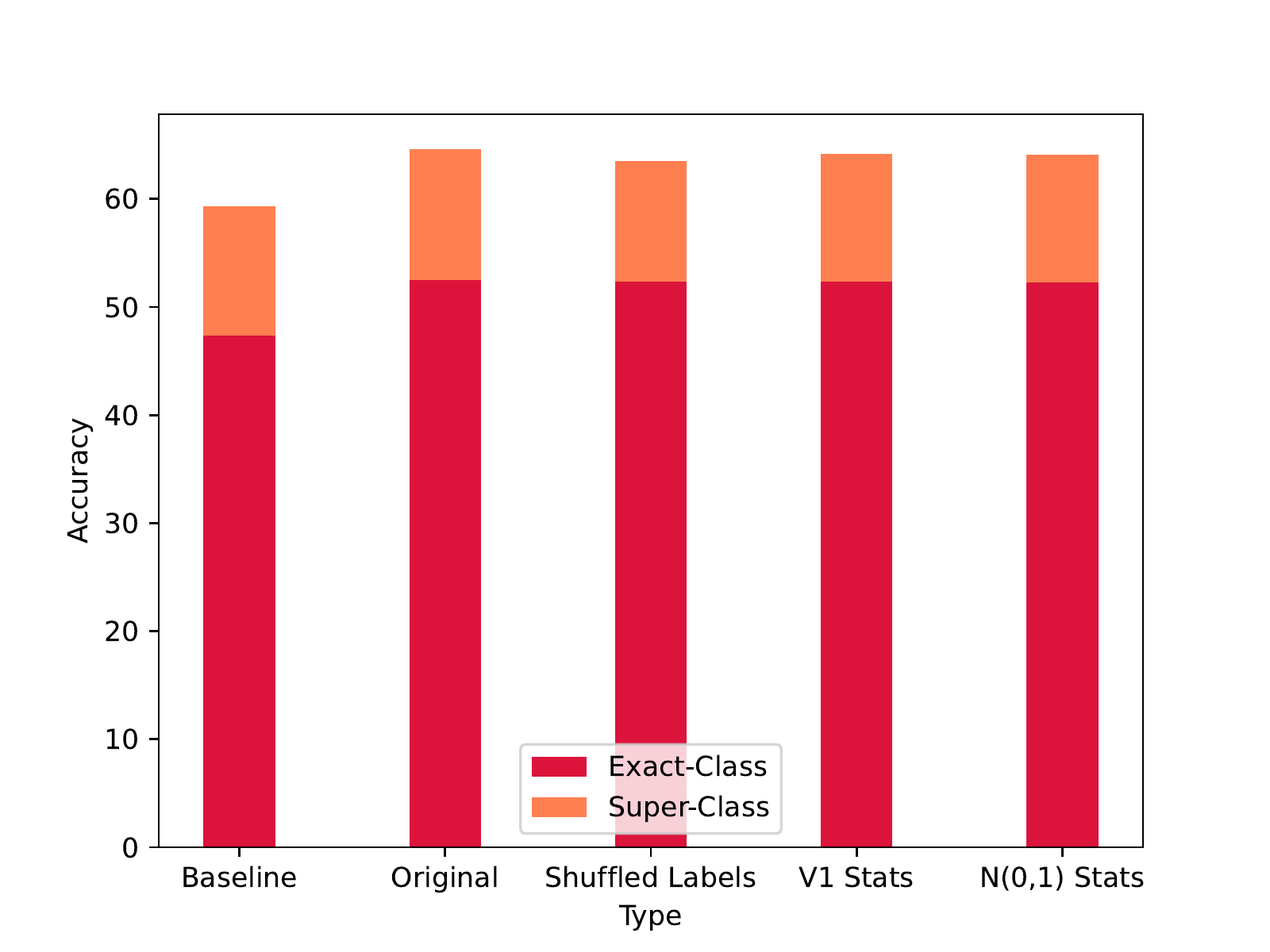}
    \includegraphics[width=0.4736\textwidth]{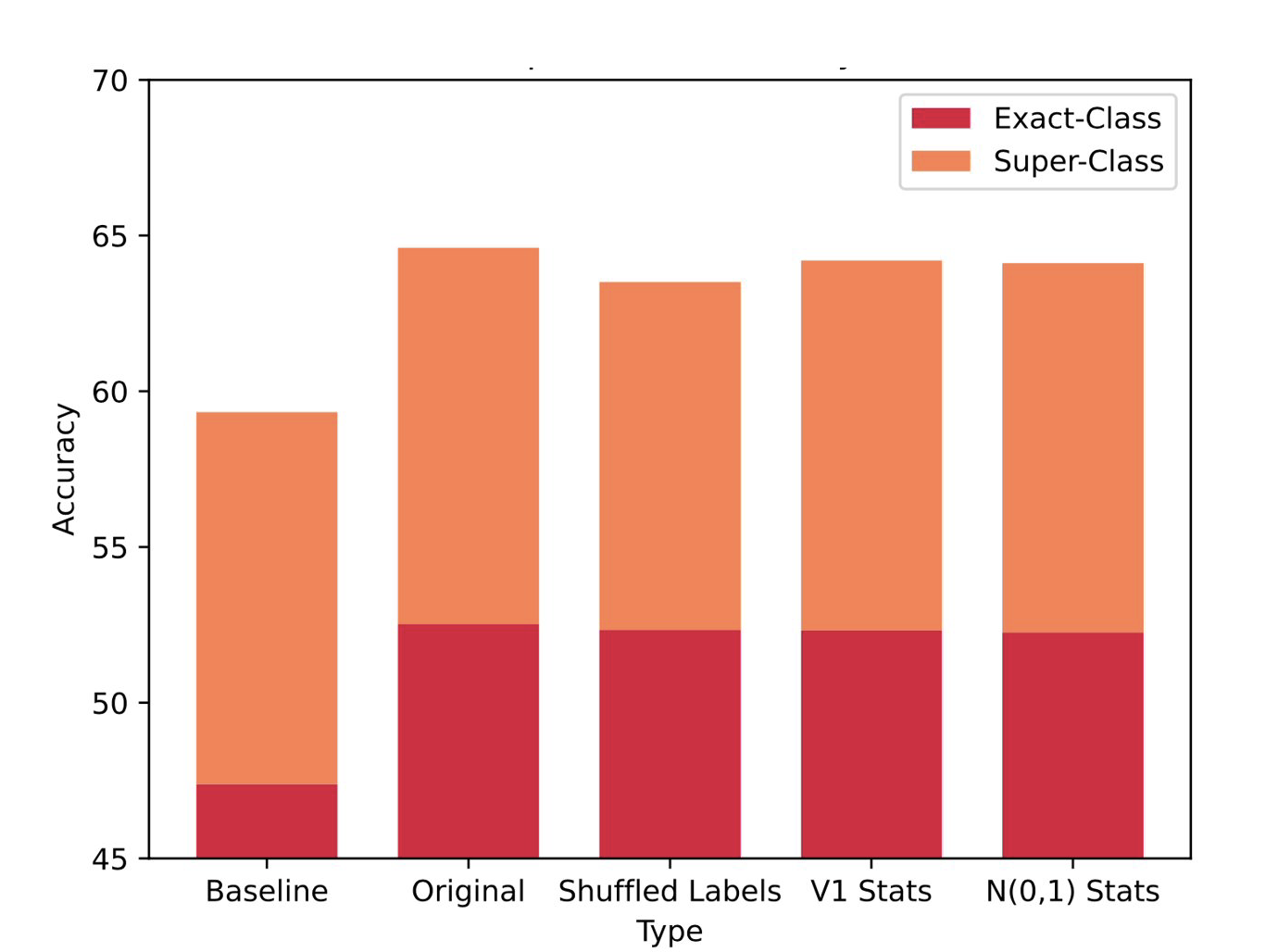}
    \caption{Super-Class Accuracy experiment on the different datasets. The best performing $\lambda$ for all datasets was $\lambda = 0.75$. The original dataset is shown to achieve better accuracy and super-class accuracy. Further, the original dataset correctly super-classifies a higher proportion of incorrectly classified images than the other datasets. In other words, although there are less misclassified images from the original dataset, it still achieves a higher proportion of correctly super-classified, but incorrectly classified images}
    \label{fig:si_all}
\end{figure}
\newpage
\subsection{Robustness to Adversarial Examples}

\begin{figure}[hbtp]
    \centering
    \includegraphics[width=0.5\textwidth]{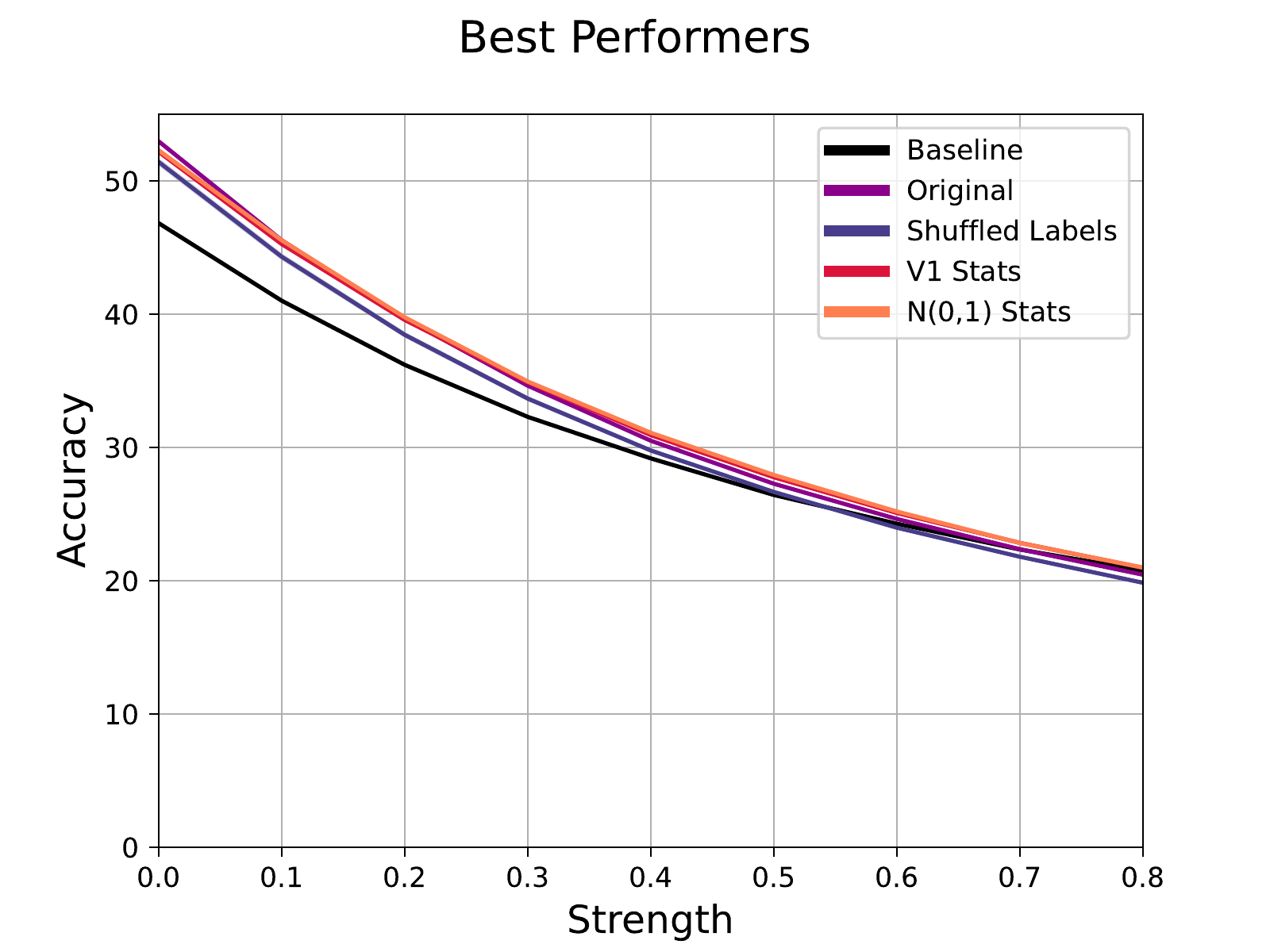}
    \caption{Robustness to Adversarial Examples experiment on the different datasets. The best performing $\lambda$ for all datasets was $\lambda = 0.75$. Note that with the ND regularizer all of the used datasets stay above the baseline until a strength of 0.5. We see the the original dataset does better at lower strengths before the standard normal dataset takes over at around 2.5.  Interestingly, as the strength increases the V1 and N(0,1) datasets do better than the shuffled labels dataset.  We suspect this is because the shuffled labels data still has proper neural recordings but now are mismatched, while the CNN is creating its own properly matched to the stimuli. Essentially working against each other.}
    \label{fig:si_best_adv}
\end{figure}

\end{document}